\documentclass[10pt,twocolumn]{article}

\usepackage[margin=1in]{geometry}
\usepackage{amsmath,amssymb}
\usepackage{graphicx}
\usepackage{booktabs}
\usepackage{float}
\usepackage{hyperref}
\usepackage{url}
\usepackage{caption}
\usepackage{subcaption}
\usepackage{microtype}
\usepackage{xcolor}
\usepackage{verbatim}
\usepackage{cite}
\usepackage{bbm}   
\usepackage{stfloats}
\usepackage{abstract}

\usepackage{authblk}
\usepackage{orcidlink} 
\title{Generating Satellite Imagery Data for Wildfire Detection through Mask-Conditioned Generative AI}

\author[1]{Valeria Martín\orcidlink{0009-0000-3668-5003}}
\author[1,2]{K. Brent Venable\orcidlink{0000-0002-1092-9759}}
\author[1]{Derek Morgan\orcidlink{0000-0003-2321-3765}}

\affil[1]{Department of Intelligent Systems and Robotics,
          University of West Florida, Pensacola, FL, USA}
\affil[2]{IHMC, Pensacola, FL, USA}

\date{}   

\begin{document}
\maketitle

\begin{abstract}
The scarcity of labeled satellite imagery remains a fundamental bottleneck
for deep-learning (DL)-based wildfire monitoring systems.  This paper
investigates whether a diffusion-based foundation model for Earth
Observation (EO), EarthSynth, can synthesize realistic post-wildfire
Sentinel-2 RGB imagery conditioned on existing burn masks, without
task-specific retraining.  Using burn masks derived from the
CalFireSeg-50 dataset \cite{martin2025sigspatial}, we design and evaluate six controlled experimental
configurations that systematically vary: (i) pipeline architecture
(mask-only full generation vs.\ inpainting with pre-fire context),
(ii) prompt engineering strategy (three hand-crafted prompts and a
VLM-generated prompt via Qwen2-VL), and (iii) a region-wise color-matching
post-processing step.  Quantitative assessment on 10 stratified test
samples uses four complementary metrics: Burn IoU, burn-region color
distance ($\Delta C_{\text{burn}}$), Darkness Contrast, and Spectral
Plausibility.  Results show that inpainting-based pipelines consistently
outperform full-tile generation across all metrics, with the structured
inpainting prompt achieving the best spatial alignment (Burn IoU\,=\,0.456)
and burn saliency (Darkness Contrast\,=\,20.44), while color matching
produces the lowest color distance ($\Delta C_{\text{burn}}$\,=\,63.22) at
the cost of reduced burn saliency.  VLM-assisted inpainting is competitive
with hand-crafted prompts.  These findings provide a foundation for
incorporating generative data augmentation into wildfire detection
pipelines.  Code and experiments are available at:
\url{https://www.kaggle.com/code/valeriamartinh/genai-all-runned}.
\end{abstract}

\section{Introduction}
\label{sec:intro}
the scarcity of high-quality labeled data remains a fundamental bottleneck for DL–based operational wildfire monitoring systems \cite{Adegun2023}.  Wildfire events are geographically sparse, seasonally concentrated, and spectrally variable depending on vegetation type, fire severity, and time since the event \cite{zahabnazouri2025burnseverity}. Although large volumes of satellite imagery are publicly available, building labeled datasets for DL-driven wildfire detection requires aligning cloud-free pre-fire and post-fire acquisitions, obtaining or generating binary masks through time-intensive manual or semi-automated processes, and tiling large satellite scenes into model-compatible patches. 

Given this labeling bottleneck, augmentation techniques can be employed to increase the effective size of available training data without requiring additional manual annotation. Data augmentation has long been used in computer vision to improve model generalization \cite{shorten2019augmentation}. Classical approaches, such as random flips, crops, color transformations, and cutouts, operate directly on existing images and introduce geometric or photometric variability without altering the underlying semantic content. In remote sensing applications, however, these transformations may offer more limited benefits, as spatial context and spectral statistics often play an important role in model performance \cite{rs15030827}.

This imbalance between imagery availability and annotation availability motivates the exploration of generative approaches capable of synthesizing additional training samples while leveraging existing masks or labels. Rather than transforming existing images, generative models aim to produce new semantically plausible examples that expand the training data. Early efforts in remote sensing image synthesis relied on Generative Adversarial Networks (GANs) \cite{DISASTERGAN}, including conditional and cycle-consistent architectures for scene generation, domain adaptation, and change detection. More recently, diffusion-based generative models have emerged as a preferred alternative for high-fidelity image synthesis. Stable Diffusion \cite{rombach} is a latent diffusion model that performs denoising in a learned lower-dimensional representation, enabling computationally efficient high-resolution image generation conditioned on text. Building on this framework, ControlNet \cite{zhang2023controlnet} introduced additional trainable conditioning branches that incorporate structured spatial signals, such as edges, depth maps, or semantic masks, into the generation process. This design enables fine-grained spatial control while keeping the pretrained diffusion backbone fixed, making it particularly suitable for scenarios where existing masks can serve as conditioning inputs.

In this work, we investigate whether a diffusion-based foundation model for EO, EarthSynth \cite{earthsynth}, can be adapted to synthesize realistic post-wildfire satellite imagery conditioned on existing burn masks. Unlike the general remote sensing categories on which EarthSynth was trained, wildfire burn scars exhibit distinct spectral characteristics (e.g., dark charcoal and ash tones), irregular spatial boundaries, and subtle gradients between fully and partially burned vegetation \cite{zahabnazouri2025burnseverity}. Adapting EarthSynth to this domain, therefore, presents a domain-specific generation challenge.

To evaluate this systematically, we design six controlled experiments that vary: (i) pipeline architecture (full-generation vs.\ inpainting), (ii) prompt strategy (three hand-crafted prompts and a VLM-generated prompt), and (iii) a color post-processing step. The experiments assess synthesis quality by comparing generated imagery to real post-fire Sentinel-2 RGB images.

The contributions of this paper are:
\begin{itemize}
  \item A generative augmentation framework for wildfire satellite imagery
        that leverages existing burn masks as conditioning inputs.
  \item The adaptation of EarthSynth to the wildfire domain without
        task-specific retraining.
  \item Six controlled experimental configurations isolating the effects of
        pipeline architecture and prompt strategy.
  \item Quantitative and qualitative evaluation of generation quality
        across burn-ratio strata.
\end{itemize}

The remainder of this paper is organized as follows.
Section~\ref{sec:background} provides the necessary background on diffusion
models, inpainting, and VLMs.  Section~\ref{sec:related} reviews related
work.  Section~\ref{sec:genai_methods} describes the experimental design and
evaluation metrics.  Section~\ref{sec:ch6_results} presents results.
Section~\ref{sec:genai_discussion} discusses implications and limitations, and
Section~\ref{sec:genai_conclusion} concludes.

\section{Background}
\label{sec:background}

\subsection{Denoising Diffusion Probabilistic Models}
\label{subsec:ddpm}

Denoising diffusion models generate new samples by learning to reverse a
gradual noising process~\cite{sohl2015noneq,ho2020ddpm}.  The core idea:
start with real data (satellite images), progressively add Gaussian noise
over many timesteps until the data become indistinguishable from pure
noise, then train a neural network (NN) to reverse this process,
iteratively removing noise to recover realistic images from random starting
points~\cite{ho2020ddpm}.

\subsubsection{Forward Process}

The forward diffusion process defines a Markov chain that slowly destroys
structure in data $\mathbf{x}_0 \sim q(\mathbf{x}_0)$ over $T$ timesteps
(typically $T=1000$).  At each step a small amount of Gaussian noise is
added:
\begin{equation}
q(\mathbf{x}_t | \mathbf{x}_{t-1}) =
\mathcal{N}\!\bigl(\mathbf{x}_t;\,\sqrt{1-\beta_t}\,\mathbf{x}_{t-1},\,
\beta_t\mathbf{I}\bigr),
\end{equation}
where $\beta_t\in(0,1)$ is a variance schedule.  A closed-form marginal
allows direct sampling at any timestep:
\begin{equation}
\mathbf{x}_t = \sqrt{\bar\alpha_t}\,\mathbf{x}_0
              + \sqrt{1-\bar\alpha_t}\,\boldsymbol\epsilon,\quad
\boldsymbol\epsilon\sim\mathcal{N}(0,\mathbf{I}),
\end{equation}
where $\alpha_t=1-\beta_t$ and $\bar\alpha_t=\prod_{s=1}^{t}\alpha_s$.

\subsubsection{Reverse Process and Training}

The reverse conditional distribution is parameterized by a NN:
\begin{equation}
p_\theta(\mathbf{x}_{t-1}|\mathbf{x}_t)=
\mathcal{N}\!\bigl(\mathbf{x}_{t-1};\,\boldsymbol\mu_\theta(\mathbf{x}_t,t),\,
\Sigma_\theta(\mathbf{x}_t,t)\bigr).
\end{equation}
Rather than predicting $\boldsymbol\mu_\theta$ directly, the network
predicts the added noise $\boldsymbol\epsilon_\theta(\mathbf{x}_t,t)$ and
is trained with the simplified objective~\cite{ho2020ddpm}:
\begin{equation}
\mathcal{L}_{\text{simple}}=
\mathbb{E}_{t,\mathbf{x}_0,\boldsymbol\epsilon}
\bigl[\|\boldsymbol\epsilon -
       \boldsymbol\epsilon_\theta(\mathbf{x}_t,t)\|^2\bigr].
\end{equation}

\subsubsection{Conditioning}

Controlled generation uses three complementary mechanisms~\cite{rombach}:
\emph{concatenation} of conditioning signals with $\mathbf{x}_t$;
\emph{cross-attention} injection of text or spatial embeddings; and
\emph{classifier-free guidance}~\cite{ho2022classifierfree}:
\begin{equation}
\tilde{\boldsymbol\epsilon}_\theta =
  \boldsymbol\epsilon_\theta(\mathbf{x}_t,\emptyset)
  + w\cdot\bigl(\boldsymbol\epsilon_\theta(\mathbf{x}_t,c)
               -\boldsymbol\epsilon_\theta(\mathbf{x}_t,\emptyset)\bigr),
\end{equation}
where $c$ is the conditioning signal and $w>1$ amplifies its influence.

\subsection{Inpainting for Controlled Synthesis}
\label{subsec:inpainting}

Inpainting reconstructs missing or occluded regions given a partially
observed image $\mathbf{x}_{\text{observed}}$ and a binary mask
$\mathbf{m}\in\{0,1\}^{H\times W}$~\cite{zhu2012cloud}.  The generative
process conditions on $\mathbf{x}_{\text{observed}}\odot\mathbf{m}$ and
$\mathbf{m}$, producing content consistent with the preserved context.

In this paper, inpainting is repurposed for \emph{controlled insertion} of
wildfire damage: the diffusion model generates post-fire texture and
coloration inside masked burn regions while preserving the surrounding
unburned landscape.  This formulation leverages existing burn masks to
synthesize post-event imagery without modifying unaffected areas.

\subsection{Generating Images from Burn Masks (EarthSynth)}
\label{subsec:earthsynth}

EarthSynth~\cite{earthsynth} is a conditional diffusion model built on
Stable Diffusion v1.5 with a ControlNet module trained on EarthSynth-180K,
a 180{,}000-sample dataset spanning diverse land-cover categories.  The
model accepts (i) a semantic mask for spatial conditioning and (ii) a text
prompt for semantic conditioning, supporting dual-control generation within
a single framework.  Its mask-based interface allows existing burn-scar
masks to serve directly as spatial conditioning inputs, while text prompts
steer spectral characteristics toward Sentinel-2-like RGB appearance.

\subsection{Vision Language Models for Prompt Generation}
\label{subsec:vlm}

Vision Language Models (VLMs) are multimodal architectures that jointly
process visual and textual information~\cite{radford2021clip,li2023blip2}.
Modern VLMs combine a visual encoder with a large language model through
cross-attention or projection layers, enabling grounded text generation
from visual inputs.  In the context of prompt-conditioned diffusion
generation, VLMs offer a scalable alternative to manual prompt engineering:
by converting visual examples into structured textual descriptions, they
can generate semantically grounded prompts that capture domain-specific
appearance without requiring human intervention for each
sample~\cite{Cheng2025RobustnessVLMArtifacts}.

\section{Related Work}
\label{sec:related}
In recent years, diffusion-based generative models have been increasingly applied to the synthesis of remote sensing imagery due to the need for labeled data augmentation and controllable scene generation. This section reviews prior work in this field, showing its progression from general-purpose satellite image generation toward spatially controllable synthesis, and motivating the selection of a diffusion model as the generative backbone for this study.

The earliest efforts to build generative foundation models for satellite imagery focused on adapting diffusion architectures to the unique characteristics of remote sensing data. DiffusionSat~\cite{khanna2024diffusionsat}, is among the first such models. Trained on multiple publicly available datasets (including fMoW and SpaceNet), DiffusionSat conditions generation on geospatial metadata (latitude, longitude, timestamp, ground sampling distance) rather than text captions, which are sparsely available for satellite images. It supports temporal generation, multispectral super-resolution, and inpainting via a 3D ControlNet extension. While DiffusionSat demonstrated that diffusion models can produce realistic satellite imagery at scale, its reliance on metadata conditioning rather than semantic masks limits direct applicability to tasks that require explicit spatial layout control, such as generating burn scars within a specified region. Subsequent work introduced other conditioning mechanisms to improve spatial and semantic controllability. CRS-Diff~\cite{tang2024crsdiff}, extends the conditioning of the generated images by supporting text, metadata, and image inputs through a multi-scale feature fusion mechanism. This multi-condition design yields improved generation quality and controllability, and the authors show its utility as a data engine for downstream tasks such as road extraction. Building on a similar motivation, GeoSynth~\cite{sastry2024geosynth}, combines global style control (via text prompts or geographic coordinates) with image-driven layout control (via OpenStreetMap tiles, SAM masks, or Canny edges). Trained on satellite imagery and OpenStreetMap data with automatically generated captions, GeoSynth exhibits strong zero-shot generalization across diverse geographic settings. Both CRS-Diff and GeoSynth advance the state of controllable satellite image synthesis. In addition, some work in this field has focused specifically on using diffusion-generated imagery to augment downstream detection and segmentation models. AeroGen~\cite{tang2024aerogen}, proposes a layout-controllable diffusion model designed for remote sensing object detection augmentation. It is the first model to support both horizontal and rotated bounding-box conditioning and includes an end-to-end data augmentation pipeline with a diversity-conditioned generator and a quality-filtering mechanism. AeroGen demonstrated notable improvements in detection performance on the DIOR, DIOR-R, and HRSC benchmarks. While this work validates the principle that diffusion-generated synthetic data can improve downstream model performance, its bounding-box-level layout conditioning is not directly transferable to pixel-level mask-conditioned tasks such as wildfire burn-scar synthesis, where irregular, non-rectangular boundaries must be preserved. Taken together, these works establish that diffusion models can generate realistic satellite imagery, that spatial conditioning improves controllability, and that synthetic data can benefit downstream remote sensing tasks. However, post-wildfire satellite imagery is a domain characterized by spectrally distinct burn scars (dark charcoal and ash tones), irregular spatial boundaries, and subtle gradients between fully and partially burned vegetation. Adapting a generative model to this domain requires a framework that (i) accepts binary segmentation masks as spatial conditioning inputs, (ii) supports text-based semantic guidance to steer spectral appearance, and (iii) is computationally accessible.

EarthSynth~\cite{earthsynth} satisfies some of these requirements. It proposes a unified diffusion framework trained on EarthSynth-180K, a dataset of 180{,}000 samples aggregated from multiple remote sensing datasets spanning diverse land-cover categories. The model supports dual conditioning via semantic masks and text prompts through a ControlNet architecture built on Stable Diffusion v1.5, enabling both spatial layout control and semantic guidance within a single framework. Moreover, EarthSynth is open-source with publicly available model weights, making it feasible to deploy within the computational constraints of our research. Its mask-based conditioning interface allows existing burn-scar masks from CalFireSeg-50 \cite{martin2025sigspatial} to serve directly as spatial conditioning inputs. In contrast, its text prompt interface provides a mechanism for steering the spectral characteristics of the generated burn regions toward Sentinel-2-like RGB appearance. These properties make EarthSynth a practical and well-suited foundation for investigating generative augmentation of post-wildfire satellite imagery.

In addition, it is important to note that VLMs enable automated description of spatial structure and spectral characteristics from overhead imagery~\cite{Cheng2025RobustnessVLMArtifacts}. In the context of prompt-conditioned diffusion generation, VLMs offer a scalable alternative to manual prompt engineering: by converting visual examples into structured textual descriptions, they can generate semantically grounded prompts that capture domain-specific appearance without requiring human intervention for each sample. This work evaluates VLM-assisted prompting using a remote-sensing-adapted Qwen2-VL model~\cite{bai2025qwen2vl} to generate per-sample diffusion prompts from composite panels of pre-fire, mask, and post-fire imagery.

The primary contribution of this paper is a controlled experimental evaluation of mask-conditioned diffusion-based generation for post-wildfire satellite imagery, leveraging the EarthSynth framework and burn masks derived from CalFireSeg-50. In addition to mask conditioning, we investigate the role of prompt design by comparing structured human-written prompts with VLM-generated prompts, enabling an analysis of how textual conditioning influences spatial alignment, spectral consistency, and visual plausibility.

\section{Methods}
\label{sec:genai_methods}
This section describes the experimental design, generative framework, and evaluation setup used to synthesize and assess synthetic wildfire imagery under mask-conditioned diffusion.

\subsection{Experimental Dataset}
\label{subsec:genai_dataset}
The experiments use CalFireSeg-50 imagery \cite{martin2025sigspatial}, specifically pre-fire RGB tiles and corresponding burn masks, to condition the generative model in the synthesis of realistic post-wildfire satellite imagery.

We filter burn masks with a burned-area fraction between 1\% and 95\%, excluding near-empty and near-saturated masks. We then select 10 stratified test samples (S00--S09) per burn-ratio bin: $<20\%$, $20$--$40\%$, $40$--$60\%$, $60$--$80\%$, and $>80\%$.

Additionally, 40 palette samples (disjoint from the test set) are used to compute color statistics of real post-wildfire imagery found in the CalFireSeg-50 dataset.

\subsection{EarthSynth Generative Framework}
\label{subsec:genai_framework}
EarthSynth is a conditional diffusion model built on Stable Diffusion v1.5 with a ControlNet module trained on EarthSynth-180K. The model accepts (i) a semantic mask for spatial conditioning and (ii) a text prompt for semantic conditioning.

Generation is performed using 35 denoising steps with the UniPC scheduler and a classifier-free guidance scale of 7.5. Images are synthesized at a resolution of $512 \times 512$ pixels and then resized to $224 \times 224$ to ensure consistency with the CalFireSeg-50 data.

\subsection{Pipeline Configurations}
\label{subsec:genai_pipelines}
We evaluate two pipeline architectures:

\textbf{Full generation pipeline (Base ControlNet):}
A Stable Diffusion ControlNet Pipeline generates the entire image from scratch, conditioned only on the binary burn mask and the text prompt. The mask is converted to a 3-channel control image.

\indent \textbf{Inpainting pipeline:}
A Stable Diffusion ControlNet Inpaint Pipeline takes the real "Before" image, its corresponding binary burn mask, and the text prompt, generating burned vegetation only inside the burned region while preserving the rest of the pre-fire image. 

\subsection{Prompt Engineering Strategies}
\label{subsec:genai_prompts}
Contrastive Language–Image Pretraining (CLIP) text conditioning is limited (77 Byte Pair Encoding (BPE) tokens) as longer prompts are truncated. We design three hand-crafted prompts within a 75-token budget:

\begin{itemize}
  \item \textbf{Prompt 1 (P1) --- Minimal:} \texttt{"satellite RGB image, wildfire burn scar, charred forest, aerial nadir view, no clouds, sharp"}.
  \item \textbf{Prompt 2 (P2) --- Structured:} \texttt{"optical satellite RGB image, nadir view, wildfire aftermath, burned area shows dark brown charcoal and ash tones, surrounding intact green forest canopy unchanged, Sentinel-2-like, sharp detail, no clouds, no smoke"}.
  \item \textbf{Prompt 3 (P3) --- Data-contextual:} This prompt is dynamically constructed from palette RGB statistics, converting mean colors of burned/intact regions into grounded color descriptors and inserting them into a structured template --> \texttt{"optical satellite RGB image, nadir view, wildfire burn scar aftermath, burned region: {insert dynamic burn description}, ash deposits, charcoal texture, intact region: {insert dynamic intact description}, forest canopy unchanged,  Sentinel-2-like, sharp detail, no clouds, no smoke"}.
\end{itemize}

All prompts share a common negative prompt: \texttt{"ground level view, eye level, forest interior, tree trunks, clouds, smoke, flames, buildings, cartoon, blurry, watermark, low resolution, artifacts, perspective distortion"}.

\subsection{VLM-Assisted Prompt Generation}
\label{subsec:genai_vlm_prompt}

We use a remote-sensing-adapted Qwen2-VL-2B-Instruct model \cite{bai2025qwen2vl} to generate prompts, from a composite panel of visual examples for each sample, that conditions the downstream diffusion model. The VLM is explicitly instructed to describe only the burned region from a top-down satellite perspective, emphasizing color (dark brown, black, charcoal), ash tone, texture, and contrast with surrounding vegetation, while avoiding any ground-level or oblique viewpoint language.

The exact template provided to the VLM is shown below:

\begin{center}
\fbox{%
  \begin{minipage}{0.9\linewidth}
  \ttfamily\footnotesize\setlength{\parskip}{3pt}
  ``These satellite images are viewed from directly above (nadir view),
  like Google Maps satellite mode. Each row shows:
  [BEFORE fire $|$ MASK of burned area $|$ REAL AFTER fire].
  The MASK shows where vegetation burned.
  REAL AFTER shows what the burned scar looks like from above ---
  TOP-DOWN satellite view, NOT a photo from inside a forest.
 
  Your task: describe ONLY what the burned region looks like in the AFTER
  image. Focus on: burn scar color (dark brown/black/charcoal), ash tone,
  texture, and contrast with surrounding forest.
 
  IMPORTANT RULES: Do NOT mention trees from ground level, forest
  interiors, or trunks. Do NOT use words like `lush', `towering', or
  `canopy from below'. Keep answer under 40 words. Output will be used
  inside a Stable Diffusion prompt starting with `optical satellite RGB
  image, nadir view, Sentinel-2-like'.
 
  Return ONLY valid JSON: \{"prompt\_body": "...", "neg\_prompt": "..."\}
  neg\_prompt must include: ground level view, eye level, forest interior,
  tree trunks, perspective distortion, clouds, smoke, flames, cartoon,
  blurry''.
  \end{minipage}%
}
\end{center}
 
The generated prompt body is constrained to a maximum of 50 tokens. This constraint is needed because, without an explicit upper bound, the model occasionally produces outputs that exceed the allowable token budget of the downstream diffusion model’s text encoder (e.g., the 77 BPE token limit in CLIP-based architectures), resulting in truncation and loss of semantic information.

In addition, each generated description is prepended with a fixed satellite-specific prefix prior to conditioning the diffusion model. This prefix explicitly anchors the generation to overhead satellite imagery. Without this constraint, we observed occasional viewpoint drift in which the diffusion model generated ground-level forest scenes when prompts described “burned forest” without explicit satellite context.

The fixed prefix used for all samples is:

\begin{center}
\texttt{"optical satellite RGB image, nadir view, Sentinel-2-like"}
\end{center}

\subsection{Experimental Design}
\label{subsec:genai_design}

Six experimental configurations (E1–E6) were designed to isolate three factors in the generative pipeline: (i) the generation strategy (full mask-to-image synthesis vs. context-preserving inpainting), (ii) the application of region-wise color matching to align generated RGB statistics with those observed in real post-fire imagery, and (iii) the prompt construction method (hand-crafted prompts vs. a VLM-generated prompt). A summary of these configurations is provided in Table~\ref{tab:genai_exp_design}.

Experiments E1 and E2 establish the baseline comparison between full generation and inpainting. In E1, the model synthesizes the entire post-fire image conditioned only on the burn mask (mask-only generation). In E2, the model performs inpainting: only the masked (burned) region is regenerated while the surrounding intact area is preserved from the pre-fire image. 

Experiments E3 and E4 add a post-synthesis, region-wise color matching step. The RGB mean and standard deviation for burned and intact regions are estimated from a separate subset of real post-fire images, kept disjoint from the test samples to prevent data leakage. E3 augments the inpainting baseline (E2) with this palette-based adjustment, whereas E4 applies the same procedure to the mask-only setting (E1). These experiments are designed to test whether enforcing alignment with dataset-derived spectral statistics leads to more realistic generated wildfire imagery. Importantly, EarthSynth is not trained on this dataset; it is used as a pretrained ControlNet backbone, and the only dataset-specific learning corresponds to the estimation of these color statistics for conditioning and post-processing.

Experiments E5 and E6 replace the hand-crafted prompts (P1–P3) with a single prompt generated by a VLM. The VLM receives representative visual examples and produces a constrained textual description of the burned region from a nadir satellite perspective. In E5, this prompt is used within the inpainting pipeline; in E6, it is used within the mask-only pipeline. These configurations isolate the effect of visually grounded prompt generation on synthesis quality.

This design enables controlled comparisons across pipeline choice (E1 vs.\ E2), color matching (E2 vs.\ E3 and E1 vs.\ E4), and prompt construction strategy (E2 vs.\ E5 and E1 vs.\ E6), while also permitting analysis of sensitivity across the three hand-crafted prompts (P1–P3).

\begin{table*}[t]
\centering
\caption{Summary of the six experimental configurations. Base = full generation from mask; Inpaint = before-image context. VLM experiments use a single VLM-generated prompt.}
\label{tab:genai_exp_design}
\begin{tabular}{lllll}
\hline
Exp. & Name & Pipeline & Color match & Prompt(s) \\
\hline
E1 & Mask-only, no context & Base & No & P1, P2, P3 \\
E2 & Inpainting baseline & Inpaint & No & P1, P2, P3 \\
E3 & Contextual inpainting & Inpaint & Yes & P1, P2, P3 \\
E4 & Contextual mask-only & Base & Yes & P1, P2, P3 \\
E5 & VLM-assisted inpainting & Inpaint & No & VLM \\
E6 & VLM-assisted mask-only & Base & No & VLM \\
\hline
\end{tabular}
\end{table*}
All code used to implement the generative pipelines, metrics, and experiments described in this section is publicly available\footnote{\url{https://www.kaggle.com/code/valeriamartinh/genai-all-runned}}.
\subsection{Evaluation Metrics for Generated Wildfire Imagery}
\label{subsec:genai_metrics}

This subsection presents the evaluation metrics used to partially assess the quality of generated wildfire imagery. The selected metrics are tailored to mask-conditioned generation and inpainting-based synthesis, and are designed to capture complementary aspects of \emph{mask-consistency} and \emph{region-level spectral realism}. Although a different range of perceptual and distribution-based metrics has been proposed for generative models~\cite{heusel2017fid}, the use of these metrics is beyond the scope of this work. We therefore focus exclusively on four metrics: Burn IoU, burn-region colour distance, darkness contrast, and spectral plausibility. These metrics adapt well-established concepts, such as IoU~\cite{jaccard1901distribution}, Euclidean colour distance~\cite{sharma2005ciede2000}, and distribution-based plausibility checks\cite{heusel2017fid}, to define task-specific evaluation criteria for mask-conditioned wildfire image generation.

Let $\mathbf{x}^{(g)} \in \mathbb{R}^{H \times W \times 3}$ denote a generated RGB image, $\mathbf{x}^{(r)}$ the corresponding real post-fire image, and $\mathbf{m} \in \{0,1\}^{H \times W}$ the binary burn mask, where $m_{ij}=1$ indicates burned pixels.

\subsubsection{Burn IoU (Mask-Consistent Burn Localisation)}

Burn IoU evaluates whether the pixels inferred as ``burn-like'' in the generated image concentrate within the ground-truth burn mask. The metric is based on the Jaccard similarity coefficient~\cite{jaccard1901distribution}, also known as IoU, which is widely used for evaluating segmentation quality. Because the generative model does not directly output a binary mask, we derive a proxy burn region from the generated image using an adaptive brightness threshold, consistent with the common visual signature that burned surfaces are darker than intact vegetation in optical imagery.

Let $G_{ij}$ denote the grayscale intensity of the generated image at pixel $(i,j)$, computed as the mean across RGB channels:
\begin{align}
G_{ij} = \frac{1}{3} \sum_{c=1}^{3} x^{(g)}_{ijc}.
\end{align}

Let $p = \frac{1}{HW} \sum_{ij} m_{ij}$ denote the burn ratio. We define a threshold $\tau$ as the $p$-th percentile of $G$, and obtain a predicted burn mask:
\begin{align}
\hat{m}_{ij} =
\begin{cases}
1 & \text{if } G_{ij} \leq \tau, \\
0 & \text{otherwise}.
\end{cases}
\end{align}

Burn IoU is then defined as the Intersection-over-Union between $\hat{\mathbf{m}}$ and $\mathbf{m}$:
\begin{align}
\text{IoU}_{\text{burn}} =
\frac{\sum_{ij} \hat{m}_{ij} m_{ij}}
{\sum_{ij} \mathbb{1}\left(\hat{m}_{ij} + m_{ij} > 0\right)}.
\end{align}

Higher values indicate that the darkest (most burn-like) pixels, under this proxy definition, overlap more strongly with the ground-truth burn extent.

\subsubsection{Burn-Region Colour Distance (Region-Level Spectral Agreement)}

When a real post-fire image $\mathbf{x}^{(r)}$ is available, spectral agreement within the burned region is quantified via the Euclidean distance between mean RGB vectors. Euclidean colour distance is a standard measure of colour dissimilarity in image quality assessment~\cite{sharma2005ciede2000}.

Let $\mathcal{B} = \{(i,j) : m_{ij}=1\}$ denote the burned region. The mean RGB vectors in the generated and real images are:
\begin{align}
\boldsymbol{\mu}^{(g)}_{\text{burn}} &= 
\frac{1}{|\mathcal{B}|} \sum_{(i,j)\in\mathcal{B}} \mathbf{x}^{(g)}_{ij}, \\
\boldsymbol{\mu}^{(r)}_{\text{burn}} &= 
\frac{1}{|\mathcal{B}|} \sum_{(i,j)\in\mathcal{B}} \mathbf{x}^{(r)}_{ij}.
\end{align}

The burn-region colour distance is:
\begin{align}
\Delta C_{\text{burn}} =
\left\| \boldsymbol{\mu}^{(g)}_{\text{burn}} - 
\boldsymbol{\mu}^{(r)}_{\text{burn}} \right\|_2.
\end{align}

Lower values indicate closer agreement between the average RGB appearance of the generated and real burned areas.

\subsubsection{Darkness Contrast (Reference-Free Luminance Separation)}

Burned areas in optical imagery typically appear darker than surrounding intact vegetation. Darkness contrast measures whether this expected \emph{relative} luminance separation holds in the generated image, without requiring a ground-truth post-fire reference.

Let $\mathcal{I} = \{(i,j) : m_{ij}=0\}$ denote the intact region. Using grayscale intensities $G_{ij}$ as defined above, we compute:
\begin{align}
\mu_{\text{burn}} &= \frac{1}{|\mathcal{B}|} \sum_{(i,j)\in\mathcal{B}} G_{ij}, \\
\mu_{\text{intact}} &= \frac{1}{|\mathcal{I}|} \sum_{(i,j)\in\mathcal{I}} G_{ij}.
\end{align}

Darkness contrast is defined as:
\begin{align}
\text{DC} = \mu_{\text{intact}} - \mu_{\text{burn}}.
\end{align}

Higher values indicate that, on average, the intact region is brighter than the burned region in the generated image, consistent with typical burn-scar appearance in optical imagery.

\subsubsection{Spectral Plausibility (Mean-Colour Consistency with Dataset Palette)}

To assess whether generated burned regions are consistent with the dataset's observed burned-area colour statistics, we compare simple RGB summaries against palette-derived reference statistics learned from a disjoint subset of real post-fire imagery. This approach draws on the broader principle of comparing generated-image statistics against reference distributions, as exemplified by the Fr\'{e}chet Inception Distance~\cite{heusel2017fid}, but operates directly in RGB space rather than in a learned feature space. 
Let $\boldsymbol{\mu}^{(p)}_{\text{burn}}$ and $\boldsymbol{\sigma}^{(p)}_{\text{burn}}$ denote the dataset-level mean and standard deviation of burned-region RGB values. For the generated image, let $\boldsymbol{\mu}^{(g)}_{\text{burn}}$ denote the mean RGB vector in the burned region. We compute the normalized deviation:
\begin{align}
\mathbf{z} = 
\frac{\left| \boldsymbol{\mu}^{(g)}_{\text{burn}} - 
\boldsymbol{\mu}^{(p)}_{\text{burn}} \right|}
{\boldsymbol{\sigma}^{(p)}_{\text{burn}} + \epsilon},
\end{align}
where $\epsilon$ prevents division by zero.

Spectral plausibility is defined as the fraction of RGB channels whose deviation lies within a fixed tolerance (e.g., $2\sigma$):
\begin{align}
\text{SP} =
\frac{1}{3} \sum_{c=1}^{3} 
\mathbb{1}\left( z_c \leq 2 \right).
\end{align}

This score ranges from 0 to 1, with higher values indicating that the generated burned-region mean colour falls closer to the dataset's burned-region mean under the chosen per-channel tolerance.

Together, these four metrics provide complementary evaluation of generated wildfire imagery by quantifying: (i) whether burn-like pixels (as inferred from relative darkness) align with the provided burn mask (Burn IoU), (ii) region-level mean-colour agreement with real post-fire imagery ($\Delta C_{\text{burn}}$), (iii) reference-free luminance separation between intact and burned regions (DC), and (iv) consistency of the generated burned-region mean colour with dataset-level burned-area statistics (SP).

\subsubsection{Quality Considerations}

Generated imagery must meet several quality criteria:

\textbf{Photorealism}: Images should be visually similar from real Sentinel-2 imagery, with realistic texture, color gradients, and spatial coherence. Obvious artifacts (unrealistic colors, blurry boundaries, repeated patterns) indicate poor generation quality.

\textbf{Spectral accuracy}: Generated RGB values should match the spectral distribution of actual burned areas. If generated images exhibit systematically different spectral properties (e.g., too dark, wrong color balance), models trained on them may fail to generalize to new, unseen data.

\textbf{Geographic diversity}: The model should generalize across diverse ecosystems, forest, shrubland, grassland, producing appropriate burned appearance for each. 

\textbf{Spatial consistency}: Generated imagery should maintain realistic spatial structure, so terrain features, roads, water bodies should remain fixed, with only vegetation characteristics changing due to fire.

Quality is evaluated using the metrics defined above and through visual 
inspection by domain experts \cite{heusel2017fid,zhang2018lpips}. Ultimately, augmented model performance provides the most meaningful quality assessment.
\section{Results}
\label{sec:ch6_results}
This section reports the results for the experimental configurations and prompt strategies explained in Section \ref{sec:genai_methods}.
\subsection{Metric Results}

\begin{figure*}[h]
  \centering
  \includegraphics[width=\linewidth]{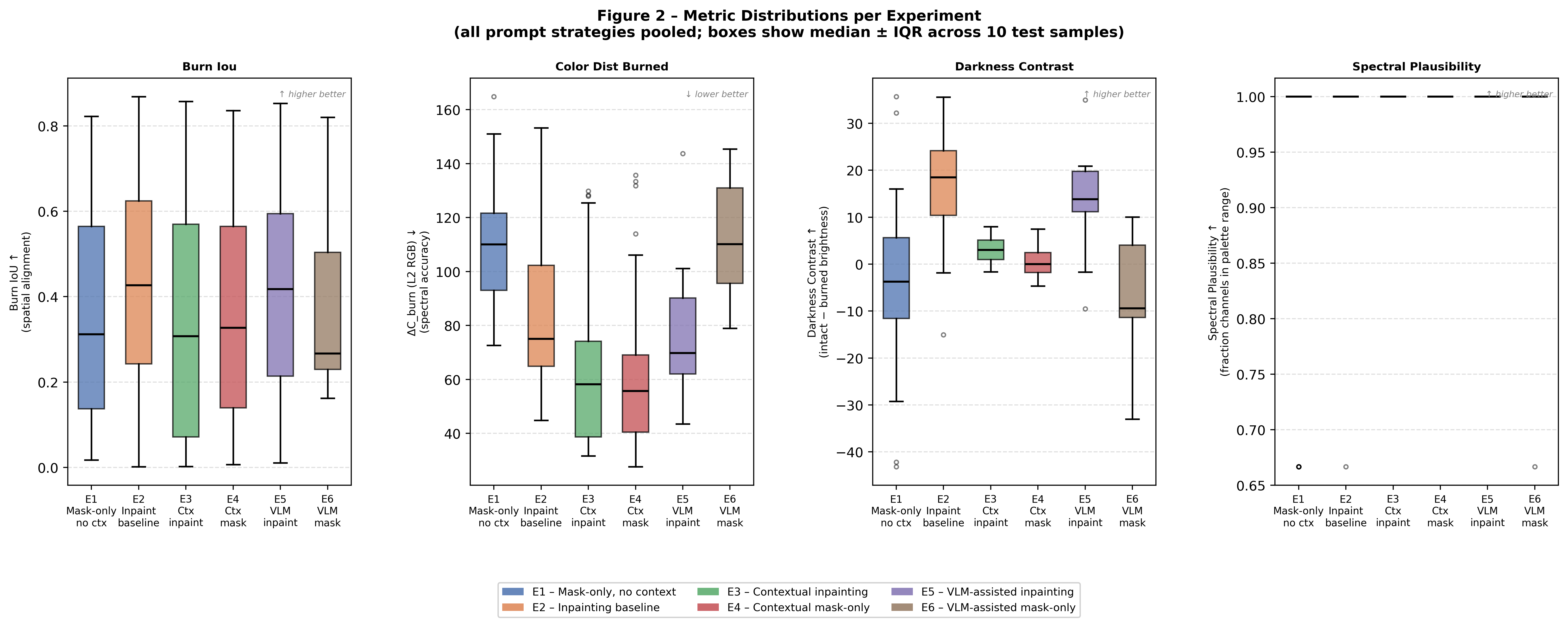}
  \caption{Metric distributions per experiment (all prompt strategies pooled). Boxes show median $\pm$ IQR across 10 test samples.}
  \label{fig:ch6_boxplots}
\end{figure*}

\begin{figure*}[t]
  \centering
  \includegraphics[width=\linewidth]{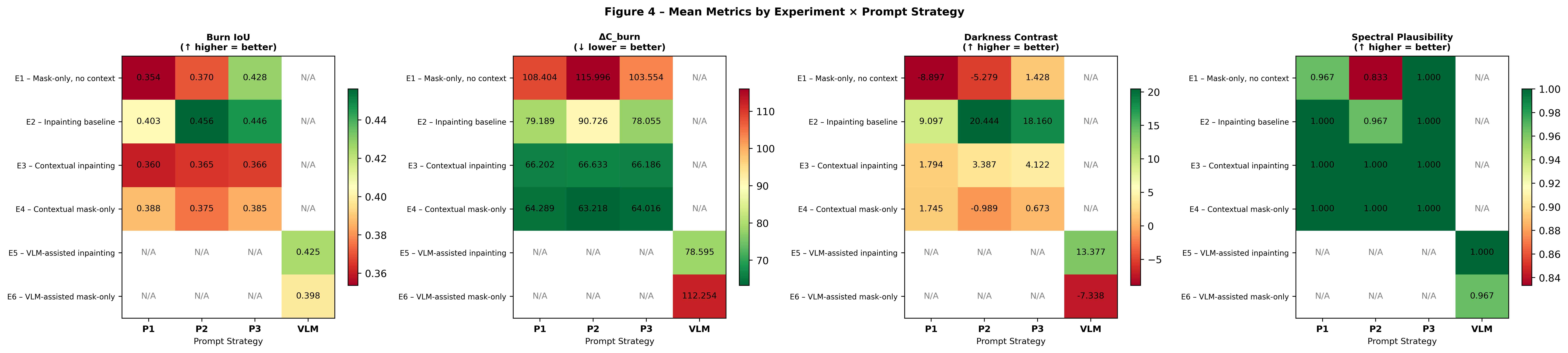}
  \caption{Mean metrics by experiment $\times$ prompt strategy.}
  \label{fig:ch6_heatmaps}
\end{figure*}

We first report the results obtained using the chosen metrics (see Subsection \ref{subsec:genai_metrics}). Mean values are computed over 10 test samples for each experiment--prompt setting (Table~\ref{tab:ch6_results}). Metric distributions aggregated at the experiment level (prompt strategies pooled) are shown in Figure~\ref{fig:ch6_boxplots}. Mean values by experiment and prompt strategy are visualized in Figure~\ref{fig:ch6_heatmaps}.

\begin{table*}[t]
\centering
\caption{Mean metric values across 10 test samples per experiment and prompt strategy}
\label{tab:ch6_results}
\begin{tabular}{llcccc}
\toprule
Exp. & Prompt & Burn IoU $\uparrow$ & $\Delta C_\text{burn}$ $\downarrow$ & Darkness Contrast $\uparrow$ & Spectral Plausibility $\uparrow$ \\
\midrule
  E1 & P1  & 0.354 & 108.40 &  -8.90 & 0.967 \\
  E1 & P2  & 0.370 & 116.00 &  -5.28 & 0.833 \\
  E1 & P3  & 0.428 & 103.55 &   1.43 & 1.000 \\
  E2 & P1  & 0.403 &  79.19 &   9.10 & 1.000 \\
  E2 & P2  & \textbf{0.456} &  90.73 &  \textbf{20.44} & 0.967 \\
  E2 & P3  & 0.446 &  78.06 &  18.16 & 1.000 \\
  E3 & P1  & 0.360 &  66.20 &   1.79 & 1.000 \\
  E3 & P2  & 0.365 &  66.63 &   3.39 & 1.000 \\
  E3 & P3  & 0.366 &  66.19 &   4.12 & 1.000 \\
  E4 & P1  & 0.388 &  64.29 &   1.75 & 1.000 \\
  E4 & P2  & 0.375 &  \textbf{63.22} &  -0.99 & 1.000 \\
  E4 & P3  & 0.385 &  64.02 &   0.67 & 1.000 \\
  E5 & VLM & 0.425 &  78.59 &  13.38 & 1.000 \\
  E6 & VLM & 0.398 & 112.25 &  -7.34 & 0.967 \\
\bottomrule
\end{tabular}
\end{table*}

Across all experiment--prompt settings, mean Burn IoU values ranged from 0.354 to 0.456 (see Table~\ref{tab:ch6_results}). The maximum mean Burn IoU was observed for E2--P2 (0.456), followed by E2--P3 (0.446) and E1--P3 (0.428). The VLM-assisted inpainting setting E5 (VLM) obtained a mean Burn IoU of 0.425, while the VLM-assisted mask-only setting E6 (VLM) obtained 0.398. The minimum mean Burn IoU occurred in E1--P1 (0.354). Within E1, Burn IoU increased from P1 (0.354) to P3 (0.428). Within E2, Burn IoU spanned 0.403--0.456 across P1--P3. Within E3, Burn IoU remained within 0.360--0.366 across prompts. Within E4, Burn IoU ranged from 0.375 to 0.388 across prompts (see Figures ~\ref{fig:ch6_boxplots} and  \ref{fig:ch6_heatmaps}). 

Mean $\Delta C_{\text{burn}}$ values ranged from 63.22 to 116.00 across all experiment--prompt settings (see Table~\ref{tab:ch6_results}). The lowest mean $\Delta C_{\text{burn}}$ was observed for E4--P2 (63.22), with E4--P1 (64.29) and E4--P3 (64.02) showing similar values. E3 produced means between 66.19 and 66.63 across prompts. E2 produced means between 78.06 and 90.73, and E5 (VLM) yielded 78.59. The largest mean $\Delta C_{\text{burn}}$ was observed in E1--P2 (116.00), with E1--P1 (108.40) and E6 (VLM) (112.25) also exhibiting larger values.

Mean Darkness Contrast values ranged from $-8.90$ to $20.44$ across all settings (Table~\ref{tab:ch6_results}). The largest mean Darkness Contrast was observed for E2--P2 (20.44), followed by E2--P3 (18.16) and E5 (VLM) (13.38). E3 produced positive means spanning 1.79--4.12 across prompts. E4 exhibited means of 1.75 (P1), $-0.99$ (P2), and 0.67 (P3). Negative means were observed for E1--P1 ($-8.90$), E1--P2 ($-5.28$), and E6 (VLM) ($-7.34$), while E1--P3 was positive (1.43).

Spectral Plausibility means were concentrated near 1.000 across most settings (Table~\ref{tab:ch6_results}). A mean of 1.000 was observed for all E3 and E4 prompt strategies, as well as E2--P1, E2--P3, and E5 (VLM). Values below 1.000 occurred in E1--P1 (0.967), E1--P2 (0.833), E2--P2 (0.967), and E6 (VLM) (0.967). 

The results of the metrics are reflected in the experiment-pooled distributions in Figure~\ref{fig:ch6_boxplots} and the per-setting heatmap in Figure~\ref{fig:ch6_heatmaps}.

\subsection{Visual Results}
Qualitative results are reported for a subset of five samples (S00, S02, S05, S06, and S08), chosen to span a wide range of burn ratios (10\%--90\%). Sample identifiers are assigned sequentially (S00–S09) during the stratified test set construction; the subset presented here is selected to provide representative coverage across burn-ratio bins rather than to reflect consecutive indexing. Table~\ref{tab:ch6_qual_summary} summarizes the dominant visual behaviors observed across experiments and prompt configurations. Descriptions in this subsection are observational; interpretation is deferred to Section~\ref{sec:genai_discussion}.

\begin{figure*}[t]
  \centering
  \includegraphics[width=\textwidth, height=0.9\textheight, keepaspectratio]{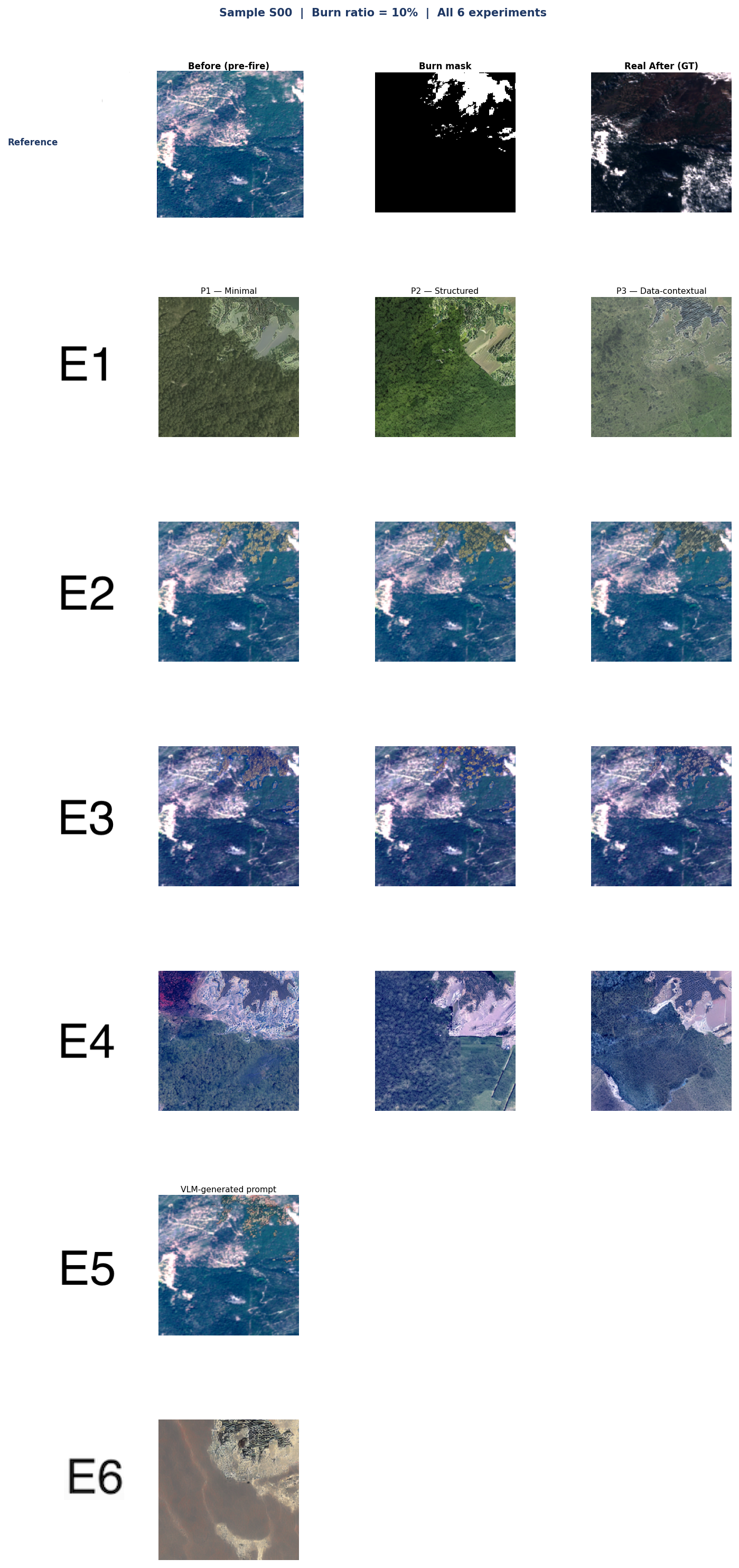}
  \centering
  \captionsetup{justification=centering}
  \caption{Visual Results for Sample S00 (burn ratio 10\%)}
  \label{fig:ch6_qual_S00}
\end{figure*}

\begin{figure*}[t]
  \centering
  \includegraphics[width=\textwidth, height=0.9\textheight, keepaspectratio]{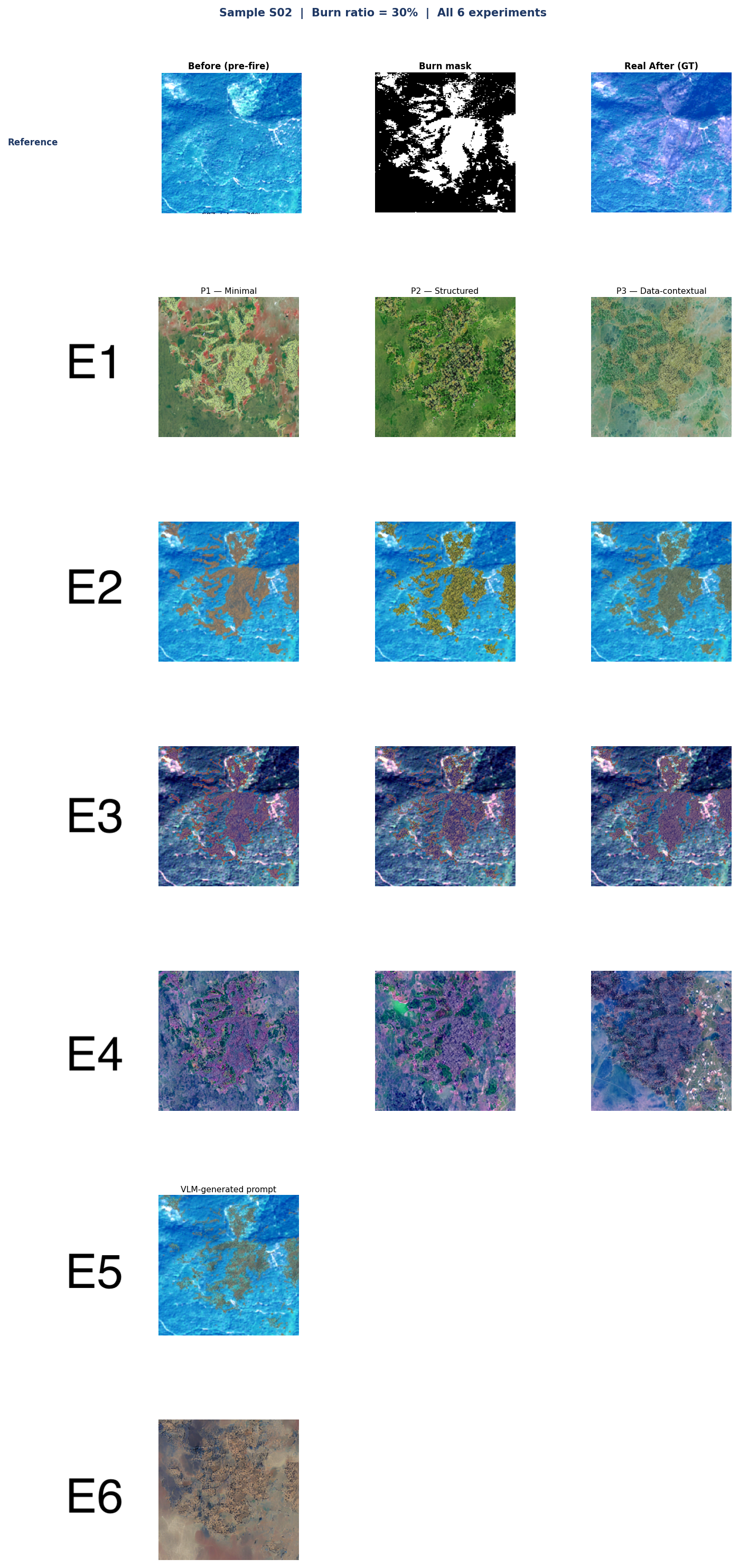}
  \captionsetup{justification=centering}
  \caption{Visual Results for Sample S02 (burn ratio 30\%)}
  \label{fig:ch6_qual_S02}
\end{figure*}

\begin{figure*}[t]
  \centering
  \includegraphics[width=\textwidth, height=0.9\textheight, keepaspectratio]{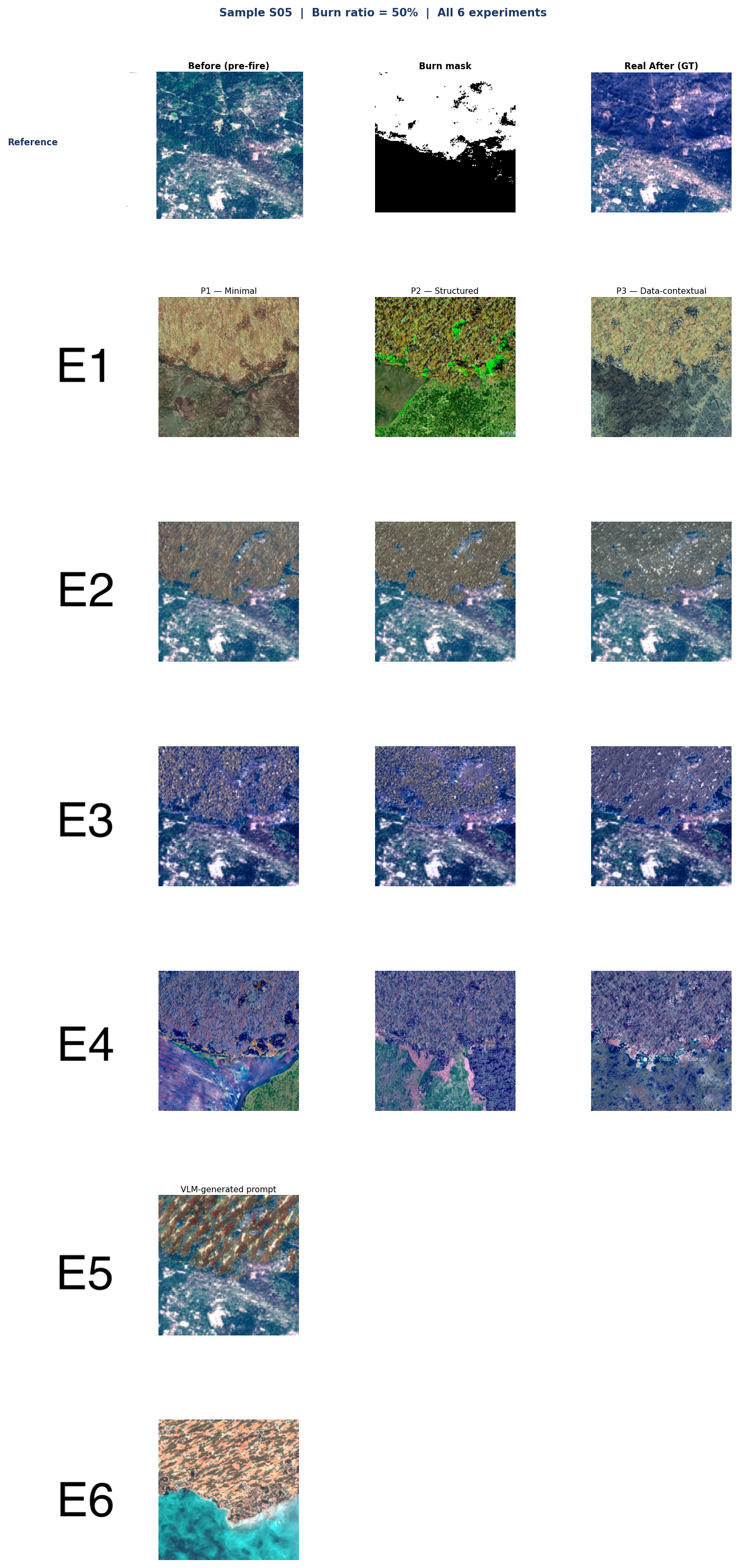}
  \captionsetup{justification=centering}
  \caption{Visual Results for Sample S05 (burn ratio 50\%)}
  \label{fig:ch6_qual_S05}
\end{figure*}

\begin{figure*}[t]
  \centering
  \includegraphics[width=\textwidth, height=0.9\textheight, keepaspectratio]{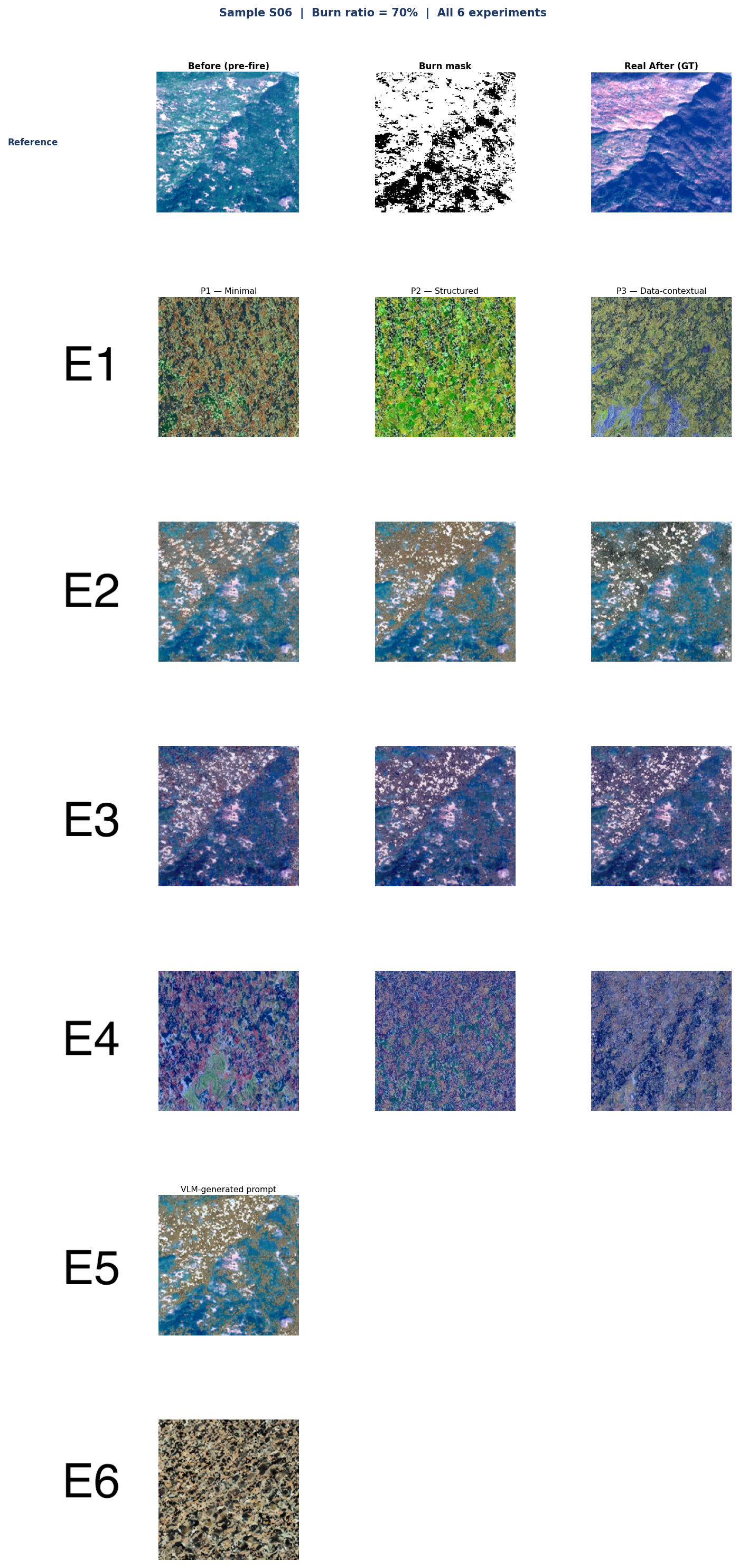}
  \captionsetup{justification=centering}
  \caption{Visual Results for Sample S06 (burn ratio 70\%)}
  \label{fig:ch6_qual_S06}
\end{figure*}

\begin{figure*}[h]
  \centering
  \includegraphics[width=\textwidth, height=0.9\textheight, keepaspectratio]{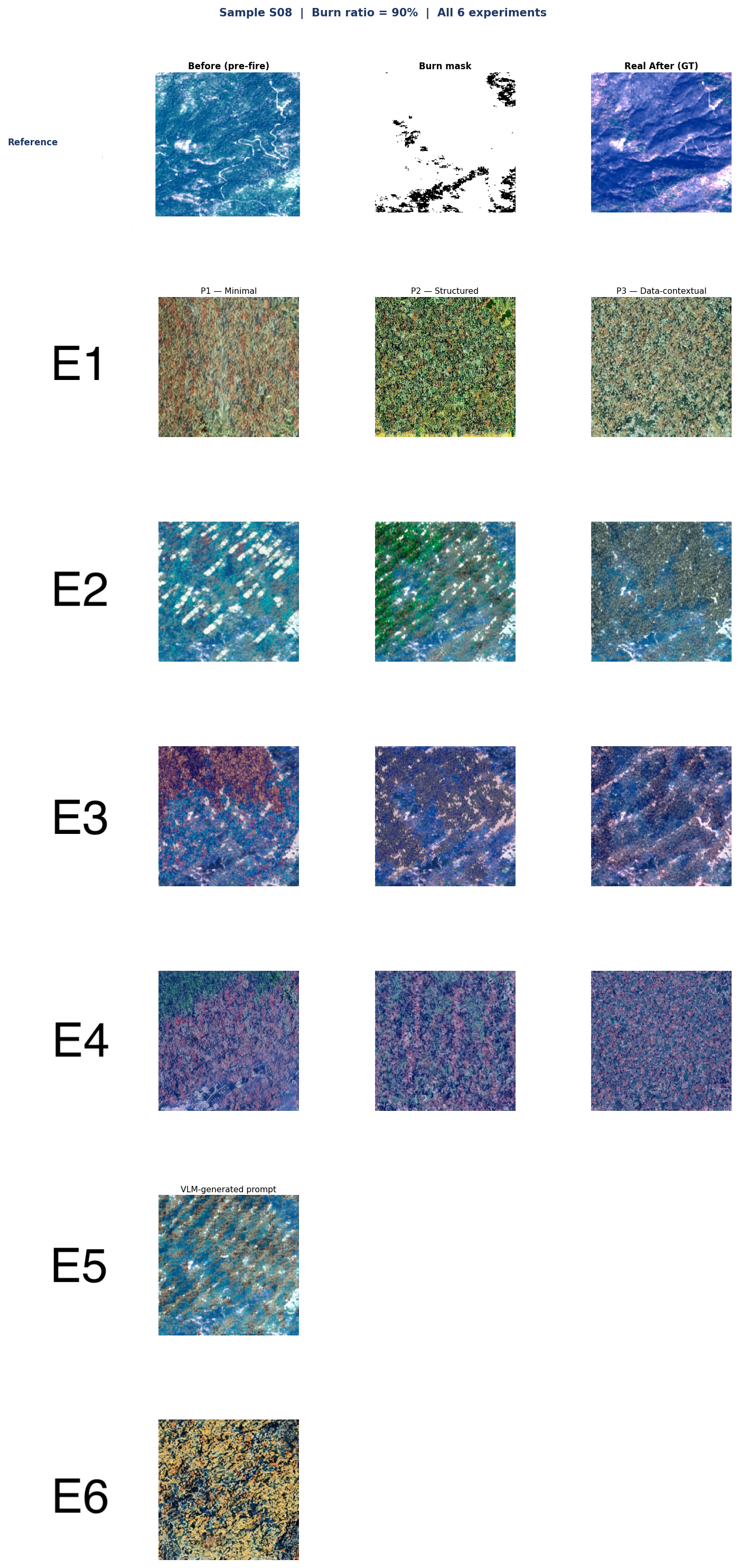}
  \captionsetup{justification=centering}
  \caption{Visual Results for Sample S08 (burn ratio 90\%)}
  \label{fig:ch6_qual_S08}
\end{figure*}

Visual results for 5 selected samples for image generation are shown in Figures~\ref{fig:ch6_qual_S00}, \ref{fig:ch6_qual_S02}, \ref{fig:ch6_qual_S05}, \ref{fig:ch6_qual_S06}, and \ref{fig:ch6_qual_S08}. They all follow the same layout: the first row shows the reference inputs (pre-fire tile, burn mask, and real post-fire tile). Rows 2, 3, 4, 5 correspond to E1--E4 and the columns correspond to the three prompt strategies (P1, P2, and P3) defined above. Rows 6 and 7 correspond to E5 and E6; in E5 the model performs VLM-assisted inpainting (conditioning on the pre-fire tile with the burn mask), while in E6 the VLM prompt is used for whole-image generation with the burn mask as condition. Across all panels, the burn mask is visualized as a binary map where white pixels indicate burned area and black pixels indicate unburned area.
In Figure \ref{fig:ch6_qual_S00}, the reference row shows a pre-fire tile with a small burned region (white) concentrated toward the upper-right, and a real post-fire tile exhibiting a darker appearance in the corresponding area.

In E1, all three prompts generate a coherent green landscape in the intact region, but the masked region does not consistently appear as a burn scar. Under P1 and P2, the masked area retains greenish tones and resembles vegetation removal rather than burning; P2 additionally introduces localized artifacts. Under P3, the masked region follows the burn mask boundary more closely and shifts toward darker tones, although the surface texture does not strongly resemble a burned pattern.

In E2, the unmasked portion of the tile remains visually consistent with the pre-fire reference, and edits are localized to the masked region. Under P1, the burn alignment is partial. Under P2, the masked region more clearly follows the burn shape but appears greener and exhibits strong color contrast relative to the surrounding vegetation. Under P3, the region appears darker and more burn-like, though noticeable color differences remain between edited and unedited areas.

\begin{table*}[t]
\centering
\caption{Qualitative summary of visual behavior by experiment. Descriptions are observational; interpretation is deferred to Section~\ref{sec:genai_discussion}.}
\label{tab:ch6_qual_summary}
\setlength{\tabcolsep}{4pt}
\renewcommand{\arraystretch}{0.95} 
\begin{tabular}{p{0.75cm} p{1.3cm} p{13cm}}
\toprule
\textbf{Exp.} & \textbf{Prompts} & \textbf{Observed visual behavior} \\
\midrule
E1 & P1--P3 &
Mask-only synthesis; prompt variants mainly shift hue (greener vs.\ muted/gray) and granularity, while burn-like appearance within the mask is inconsistent across samples. \\

E2 & P1--P3 &
Edits are in burned pixels and appear as brown/gray overlays or speckled textures. Prompt variants primarily affect tint, darkness, and fine-scale texture density within the masked region, and boundary color discontinuities can be visible. \\

E3 & P1--P3 &
Masked regions tend to show smoother blending and more integration with surrounding context than E2. Prompt variants modulate saturation and mottling. \\

E4 & P1--P3 &
Full-image generation produces appearance changes across the entire tile. Prompt variants can induce global color casts (e.g., purple/blue shifts) and sample-dependent land-cover textures; burn-like coloration may be present but is not consistent. \\

E5 & VLM &
VLM-assisted inpainting preserves the pre-fire tile outside the mask; masked edits can introduce distinctive patterns (e.g., clustered speckles or stripe-like textures). Adherence to mask extent varies by sample. \\

E6 & VLM &
Outputs often diverge from forested satellite appearance and show sample-dependent global palettes/textures; visual correspondence to real data is weak. \\
\bottomrule
\end{tabular}
\end{table*}

In E3, the pre-fire context outside the mask is preserved, and the burned region appears more coherently integrated with surrounding texture. P1 produces a consistent darkening within the mask. P2 introduces small green patches within the burned region. P3 yields the strongest mask-aligned darkening and the most visually consistent texture among the three prompts.

In E4, the entire tile is synthesized. Portions of the left side resemble a forested scene, but the global land-cover appearance varies across prompts. Burn-like coloration is present, yet areas around and beyond the mask do not consistently resemble a forested satellite image. P3 produces comparatively more muted tones, while P1 introduces stronger reddish coloration.

In E5, the masked-region edits are localized but exhibit a color mismatch relative to the surrounding unmasked vegetation.

In E6, the entire tile is synthesized and does not visually resemble a forested satellite image, diverging from both the pre-fire reference and the real post-fire tile.
\label{subsec:ch6_visual_results}

For sample S02 (see Figure \ref{fig:ch6_qual_S02}), the pre-fire tile shows a dense, uniformly green forested scene, while the real post-fire tile exhibits a clear burn-scar appearance, making this sample a strong visual reference pair. The burn mask is spatially scattered across the tile but is most concentrated near the center.

In E1, the generated tiles include reddish tones suggestive of active fire in some regions, but the surrounding vegetation does not consistently resemble a tree-like forest texture. Under P1, green vegetation persists within the masked region. Under P2, the masked region appears more altered and includes darker burn-like patches, though the surrounding forest texture remains visually inconsistent. Under P3, coloration within the masked region is more visually plausible, but large portions of the tile away from the mask lack forest-like structure.

In E2, inpainting shows edits that appear visually reasonable in color, with P3 producing the most coherent burn-like coloration among the three prompts. Across P1--P3, however, the synthesized burned textures do not strongly follow the large-scale spatial structure of the scene (e.g., terrain-aligned patterns), and the resulting burned-region appearance is primarily texture- and color-driven.

In E3, outputs are similar across P1--P3 and appear visually consistent in both tone and integration with the surrounding context. The burned-region appearance remains mask-localized, with limited prompt-dependent variation. Compared to E2, the burned-region color transitions appear less visually discontinuous. In this sample, the background pre-fire tile also appears darker, despite the use of inpainting.

In E4, P1 and P2 show prominent purple-toned global shifts that reduce the forest-like appearance of the scene, with particularly bright and saturated hues in some regions. Under P2, surrounding areas appear somewhat more coherent than P1 but still include strong color casting. Under P3, the masked region appears more burn-like, while the surrounding land-cover shows increased diversity in texture and coloration.

In E5, the VLM-generated prompt produces an output that is visually close to the strongest inpainting results in this sample, with coherent burned-region coloration and limited additional artifacts.

In E6, the output shows some alignment between the mask and the placement of altered textures, but the global appearance and surrounding regions do not resemble a forested satellite scene.

In Figure \ref{fig:ch6_qual_S05}, the burn mask covers most of the upper portion of the tile. The pre-fire reference shows a green forested scene, while the real post-fire tile exhibits clear burn-scar hues in the corresponding region, providing a strong visual contrast between intact and burned areas.

In E1, outputs vary across prompts, but the burned region and surrounding areas do not consistently exhibit forest-like structure. Under P1, the masked region does not resemble a burned forest and the surrounding vegetation remains non-forest-like, although the global palette is not severely mismatched. Under P2, the tile appears highly saturated and green, including within the masked region. Under P3, the intact (unmasked) region appears more forest-like in texture, while the masked region adopts burn-like coloration.

In E2, the three prompts produce visually similar outputs, with differences primarily expressed through small changes in overall darkness. Among the three, P3 yields the most forest-like intact appearance.

In E3, P1 and P2 produce the clearest contrast between the intact green region and the burned region, yielding a burned-area texture that appears visually consistent with the surrounding vegetation. Under P3, the burned region remains present, but the synthesized surface texture appears less forest-like.

In E4, prompt-dependent performance is pronounced. Under P1, the unburned region includes unrealistic color tones that reduce forest-like appearance. Under P2, the tile exhibits a visually strong burn/intact contrast and the surrounding vegetation appears more coherent. Under P3, the output is broadly similar in quality to P2, but small artifacts appear near the burn-mask boundary that reduce local realism.

In E5, the burned-region coloration appears inconsistent with the surrounding intact forest, and the resulting output doesn't show a coherent post-fire texture transition.

In E6, the unburned region resembles a large water body (lake- or ocean-like), while the burned region adopts burn-like hues but lacks forest-like structure. Despite these inconsistencies, this output is the only case in which E6 generates a visual that resembles a full satellite-like scene at the tile level.

For sample S06 (see Figure \ref{fig:ch6_qual_S06}), the burn mask covers the majority of the tile (approximately 70\%), leaving only small scattered intact regions. The pre-fire tile is predominantly green with some non-forested patches, while the real post-fire tile shows a strong burn-scar appearance, including very dark tones and localized reddish hues.

In E1, prompt behavior varies. Under P1, the overall result is moderately coherent but lacks realistic burn-scar texture. Under P2, vegetation remains strongly green across large portions of the tile, inconsistent with the dominant burned area. Under P3, the masked region appears largely as healthy vegetation and does not visually resemble a burn scar.

In E2, outputs are similar across prompts, with P3 producing the most burn-like coloration among the three. Across P1--P3, the burned-region appearance is primarily expressed through color and texture and does not exhibit clear alignment with large-scale scene structure; P2 in particular introduces strong brown tones that appear visually inconsistent with the surrounding context.

In E3, the generated colors are more consistent with the surrounding pre-fire scene. However, because the burned area dominates the tile, the output exhibits reduced visual variation across the masked region and appears less spatially structured relative to the underlying scene.

In E4, P3 produces the most burn-like appearance among its prompts and yields a visually plausible burned texture over the masked region. P2 exhibits color tones that are broadly compatible with burn hues but remains visually unrealistic in overall land-cover appearance. P1 includes some reddish coloration but does not yield a realistic burned-forest texture.

In E5, the output is broadly similar in appearance to E2, but includes brighter vegetation within the masked region; the vegetation does not appear fully healthy, yet the burn-scar texture remains weak.

In E6, the synthesized tile does not resemble a forested satellite image and diverges strongly in global texture and land-cover appearance, remaining the lowest-quality result in this sample.

 
Finally, for sample S08, the burn mask covers nearly the entire tile (approximately 90\%), leaving only small intact regions (see Figure \ref{fig:ch6_qual_S08}). The pre-fire and post-fire references show a strong visual contrast, with the post-fire tile exhibiting widespread burn-scar appearance across most of the scene.

In E1, outputs vary in texture and burn expression across prompts. Under P1, the scene appears dominated by widespread burning across much of the tile. Under P2, the texture appears highly grainy and lacks forest-like structure. Under P3, the output remains grainy but is comparatively closer to a forest-like appearance, with burn-like coloration distributed broadly.

In E2, prompt behavior differs strongly. Under P1, the masked region resembles cloud-like overlays rather than burn-scar texture. Under P2, green vegetation persists within the masked region in addition to cloud-like structures. Under P3, the output appears substantially more coherent, producing a darker, burn-like texture that is visually more consistent with the post-fire reference.

In E3, the masked-region extent is not fully expressed under P1, with the burn appearance covering less area than indicated by the mask. Similar partial mask-following is visible under P2 and P3, although these two outputs appear more visually consistent with satellite-like texture and coloration overall.

In E4, outputs are broadly coherent across prompts. Under P1, some regions within the burned area retain greener vegetation tones. Under P3, the output is the strongest among the three prompts, producing a more uniformly burn-like appearance across the tile, although the overall scene remains only moderately realistic in forest texture.

In E5, the output shows reduced adherence to the burn mask extent and includes stripe-like burn patterns that appear visually plausible in texture, even though the spatial placement does not consistently match the mask.

In E6, the synthesized output diverges strongly in texture and appearance.
\clearpage
\clearpage
\clearpage

\clearpage

\clearpage

\clearpage
\section{Discussion}
\label{sec:genai_discussion}

This section synthesizes the quantitative results (Table~\ref{tab:ch6_results}, Figures~\ref{fig:ch6_boxplots}--\ref{fig:ch6_heatmaps}) with the qualitative observations from the representative visual subset (Figures~\ref{fig:ch6_qual_S00}--\ref{fig:ch6_qual_S08}). The metrics capture complementary aspects of generation quality, but the visual panels are essential for understanding how scores are achieved (e.g., whether higher Burn IoU corresponds to a realistic burn scar or simply to a darker region that crosses the adaptive threshold used by the metric).

\paragraph{Pipeline choice (inpainting vs.\ mask-only generation).}
The most consistent pattern across both metrics and visuals is the advantage of context-preserving inpainting over full tile generation. Under the same prompts, the inpainting baseline (E2) improves Burn IoU (0.403--0.456 vs.\ 0.354--0.428 in E1), reduces $\Delta C_{\text{burn}}$ (78.06--90.73 vs.\ 103.55--116.00), and yields substantially higher Darkness Contrast (9.10--20.44 vs.\ $-8.90$--1.43) (Table~\ref{tab:ch6_results}). This separation is also apparent in the pooled distributions (Figure~\ref{fig:ch6_boxplots}), where E2 concentrates at higher Burn IoU and higher Darkness Contrast than E1.

The qualitative panels explain why these differences arise. In E2, the pre-fire image anchors the intact region, and synthesis is constrained to the masked area. Visually, this produces localized darkening and mottling inside the mask for multiple samples (e.g., S00 and S05), even when color discontinuities at the boundary remain. In contrast, mask-only generation (E1) must synthesize the entire tile from scratch from a binary mask and text alone. Across samples, E1 frequently produces globally inconsistent land-cover textures (often not forest-like) and masked regions that remain greenish or resemble generic texture fills rather than post-fire damage (e.g., S00 and S02). These behaviors are directly reflected in negative Darkness Contrast for E1--P1 and E1--P2 ($-8.90$ and $-5.28$), indicating that the predicted burned region can be \emph{brighter} than the intact region on average, which is incompatible with the typical darkening observed in real post-wildfire imagery.

A similar context dependence appears in the VLM settings, where E5 (VLM-assisted inpainting) achieves Burn IoU 0.425 and Darkness Contrast 13.38, whereas E6 (VLM mask-only) attains Burn IoU 0.398 and a negative Darkness Contrast ($-7.34$). Visually, E6 most often fails to resemble a forested satellite scene and instead exhibits sample-dependent global textures and palettes that do not seem realistic (consistent with its large $\Delta C_{\text{burn}}$ of 112.25). In short, automated prompting alone does not overcome the underconstraint of mask-only generation; providing spatial context remains the dominant factor.

\paragraph{Prompt strategy effects (P1--P3) and how they appear in the panels.}
Prompt sensitivity is strongest when the generation problem is least constrained. In E1, Burn IoU increases from 0.354 (P1) to 0.428 (P3), and Darkness Contrast shifts from strongly negative ($-8.90$) to positive (1.43) (Table~\ref{tab:ch6_results}). This trend matches visual behavior in S00 and S02: P3 more often produces darker mask-aligned regions than P1/P2, whereas P1 and P2 frequently preserve green tones inside the mask or introduce artifacts. However, the panels also clarify a limitation of interpreting these improvements in isolation: E1 can achieve higher Burn IoU through darker regions that satisfy the thresholding proxy without necessarily producing convincing burn-scar texture or preserving forest-like structure elsewhere in the tile.

In E2, prompt effects remain but shift in emphasis. The structured prompt (P2) yields the highest Burn IoU (0.456) and the strongest Darkness Contrast (20.44), while the data-contextual prompt (P3) yields the lowest $\Delta C_{\text{burn}}$ within E2 (78.06). This split is consistent with the visuals: P2 often produces high-contrast edits that are easier to segment as ``burned'' by the adaptive threshold used for Burn IoU (raising both IoU and Darkness Contrast), whereas P3 often pushes the burn-region colors closer to the real post-fire palette while still exhibiting boundary mismatch against the unchanged pre-fire region (e.g., S00 and S05). High-burn cases further emphasize why qualitative review matters: in S08, some outputs show incomplete mask expression or cloud-like structures inside the mask under certain prompt settings, illustrating that a strong darkness signal can inflate Darkness Contrast even when the appearance is not clearly burn-scar-like.

\paragraph{Color matching (E3/E4) and the trade-off between palette fidelity and burn saliency.}
Region-wise color matching improves burn-region color similarity, producing the best $\Delta C_{\text{burn}}$ values in the study (E4: 63.22--64.29; E3: 66.19--66.63) (Table~\ref{tab:ch6_results}). This is an expected outcome of explicitly aligning the generated burned-region RGB statistics to dataset-derived palette statistics. At the same time, both the metrics and the visuals indicate that color matching compresses contrast and reduces prompt-driven variation. Darkness Contrast is smaller under color matching (E3: 1.79--4.12; E4 includes near-zero or negative contrast such as $-0.99$), consistent with visually ``flattened'' separation between burned and intact pixels and softer boundary transitions in several samples. In addition, E3 shows minimal spread across prompts (Burn IoU 0.360--0.366; $\Delta C_{\text{burn}}$ varying by less than 0.5), mirroring the qualitative impression that outputs become more similar across P1--P3 once the post-correction changes spectral appearance.

These results suggest that color matching is effective for enforcing palette plausibility (as captured by $\Delta C_{\text{burn}}$), but it can reduce burn-scar strenght (as captured by Darkness Contrast) and reduce the influence of prompt wording. For augmentation use cases, this implies that color matching may be most useful when spectral consistency is prioritized.

\paragraph{VLM-assisted prompts (E5/E6) in the context of the hand-crafted prompts.}
E5 demonstrates that VLM-generated prompts can be competitive with hand-crafted prompting in the inpainting setting. Quantitatively, E5 (Burn IoU 0.425; $\Delta C_{\text{burn}}$ 78.59; Darkness Contrast 13.38) falls within the range of E2 outcomes and outperforms weaker hand-crafted settings, though it does not surpass the best E2 configuration (E2--P2) on Burn IoU and Darkness Contrast (Table~\ref{tab:ch6_results}). Visually, E5 often produces localized mask edits similar to E2/E3, but the panels also show distinctive patterned textures (e.g., mottling, banding, or stripe-like structure) and occasional reduced adherence to mask extent (notably in S08). In the mask-only setting, E6 remains both qualitatively and quantitatively weak, reinforcing that VLM prompting does not resolve the underconstraint of full-tile synthesis from a binary mask alone.

\paragraph{Interpreting the metrics through the qualitative panels.}
The combined evidence highlights why this work uses both quantitative and qualitative evaluation. Burn IoU captures mask alignment only through a proxy (thresholding a ``dark'' region) and can therefore improve when prompts or post-processing produce sufficiently dark pixels inside the mask even if the resulting texture is unrealistic. $\Delta C_{\text{burn}}$ captures mean color similarity in the burned region but does not indicate whether the burn pattern is spatially structured or whether boundaries transition plausibly into the surrounding context; this is evident in color-matched settings that achieve very low $\Delta C_{\text{burn}}$ while exhibiting reduced contrast. Darkness Contrast captures saliency but cannot distinguish realistic burn-scar darkening from arbitrary dark artifacts. Spectral Plausibility is high for most settings and is therefore interpreted as a coarse validity check. The visual panels are therefore critical for distinguishing ``metric-improving'' changes from qualitatively plausible post-fire structure, particularly in high burn-ratio cases where strict mask-following becomes more challenging.

\paragraph{Practical Considerations} From an augmentation perspective, the strongest candidates for producing usable image--mask pairs are the inpainting-based configurations (E2, E3 and, in some cases, E5), which more consistently preserve intact context and confine changes to the burn region. Even in these settings, the qualitative panels show boundary discontinuities and occasional partial mask expression at high burn ratios (e.g., S08), indicating that downstream use would likely benefit from additional filtering (e.g., rejecting samples), and potentially from training-time robustness to modest label noise.

In addition, the results emphasize a practical workflow point: metric summaries are informative for screening and comparison, but qualitative inspection is needed for generative augmentation in remote sensing, because visually implausible artifacts (cloud-like overlays, global recoloring, non-forest textures) may not be reliably rejected by a scalar metric. The joint use of prompt-controlled experimentation, complementary metrics, and targeted qualitative review provides a more robust basis for selecting augmentation configurations for downstream model training.

\section{Limitations and Future Work}
\label{subsec:genai_limitations}

This study has limitations that should be considered when interpreting the results and that motivate directions for future research.

First, the evaluation is conducted on only 10 stratified test samples. While stratification across burn-ratio bins ensures coverage of different spatial configurations, the small sample size limits the statistical power of cross-experiment comparisons and means that individual outlier samples can disproportionately affect mean metric values. Future work should scale the evaluation to the full CalFireSeg-50 test set to obtain more robust estimates of metric distributions and to enable testing across experimental configurations.

Moreover, this study evaluates generation quality in isolation and does not assess downstream utility. The central motivation for generative augmentation is to improve segmentation model performance under data scarcity. Future work should train DL segmentation architectures on mixed datasets of real and generated imagery and measure whether synthetic samples improve, degrade, or leave unchanged the segmentation IoU, Dice, and boundary accuracy. This downstream evaluation is important to determine whether the generation quality observed here is sufficient for practical augmentation.

In addition, EarthSynth was used without any wildfire-specific fine-tuning. The model was trained on EarthSynth-180K, which spans diverse land-cover categories but does not include post-wildfire burn-scar imagery. This domain gap likely explains the spectral inconsistencies observed across experiments, particularly the tendency to generate vegetation-like textures within masked burn regions. Fine-tuning the ControlNet module on a subset of real post-wildfire image--mask pairs from CalFireSeg-50 could reduce this domain gap and improve both spatial alignment and spectral realism. Future work could also investigate parameter-efficient fine-tuning strategies (e.g., Low-Rank Adaptation (LoRA)) to minimize the computational overhead of domain adaptation.

Another limitation is that the experiments are restricted to 3-band RGB imagery at 10\,m resolution (Sentinel-2 bands B4, B3, B2). Real wildfire detection workflows routinely NIR and SWIR bands, which provide stronger spectral contrast between burned and unburned vegetation than visible bands alone. Extending the generative pipeline to multispectral outputs would increase the utility of synthetic data for downstream models that rely on vegetation indices such as NBR or NDVI. 

Furthermore, the Burn IoU metric relies on a brightness-based thresholding proxy rather than a learned segmentation model. This proxy assumes that burned regions in the generated image will be darker than intact regions, which holds for most post-wildfire imagery but may fail for certain land-cover types (e.g., bare soil or urban surfaces that are naturally dark). Future work could apply a pretrained burn-scar segmentation model to the generated imagery and compute IoU between the predicted mask and the conditioning mask. 

For VLM prompting, future work could compare multiple VLM architectures, evaluate the effect of instruction template variations, and explore chain-of-thought or few-shot prompting strategies.

Finally, all experiments use a fixed set of generation hyperparameters (35 denoising steps, guidance scale 7.5, UniPC scheduler). The sensitivity of the results to these parameters was not explored.

\section{Conclusion}
\label{sec:genai_conclusion}

This work studied whether a diffusion-based generative foundation model for EO could be adapted to synthesize realistic post-wildfire satellite imagery without task-specific retraining. Using EarthSynth conditioned on CalFireSeg-50 burn masks, we designed and evaluated six experimental configurations that systematically varied pipeline architecture (full generation vs.\ inpainting), prompt strategy (three hand-crafted prompts and a VLM-generated prompt), and a color-matching post-processing step.

Four key findings emerged from the evaluation. First, inpainting with pre-fire spatial context consistently outperformed full mask-only generation across all metrics and in the qualitative assessment as well. Inpainting pipelines yield higher spatial alignment (Burn IoU up to 0.456 vs.\ 0.428), better color fidelity, and physically plausible burn saliency. This result could show that providing the diffusion model with real spatial context improves generation quality, particularly for a domain like post-wildfire imagery where the model has not been explicitly trained. Second, prompt specificity has a higher effect when spatial context is absent: in the mask-only pipeline, Burn IoU improved from the minimal prompt to the data-contextual prompt. In the inpainting pipeline, prompt effects were still measurable but less dominant, with the structured prompt (P2) achieving the best spatial alignment and the data-contextual prompt (P3) yielding the closest color match. Third, region-wise color matching improved mean color fidelity ($\Delta C_{\text{burn}}$) but degraded spatial accuracy (Burn IoU) and burn saliency (Darkness Contrast), revealing a trade-off between spectral correction and spatial coherence. Fourth, VLM-assisted prompting produced competitive results in the inpainting setting.

These findings show that mask-conditioned diffusion generation can produce post-wildfire satellite imagery with partial spatial and spectral fidelity.
\bibliographystyle{plain}
\bibliography{bibliography}
 
\end{document}